\documentclass[sigconf]{acmart}
\AtBeginDocument{%
  }

\setcopyright{acmlicensed}
\copyrightyear{2024}
\acmYear{2024}
\acmDOI{XXXXXXX.XXXXXXX}

\acmConference[eBRAIN lab]{}{
  2024}{NYUAD}
\acmISBN{978-1-4503-XXXX-X/18/06}

\usepackage{wrapfig}
\usepackage{tcolorbox}

\usepackage{caption}
\usepackage{adjustbox}
\usepackage{tabularx}
\usepackage{hyperref}
\begin{document}

\title{MedAide: Leveraging Large Language Models for On-Premise Medical Assistance on Edge Devices}


\author{Abdul Basit\textsuperscript{1}, Khizar Hussain\textsuperscript{2}, Muhammad Abdullah Hanif\textsuperscript{1}, Muhammad Shafique\textsuperscript{1}\\
{\small\textsuperscript{1}Division of Engineering, New Your University Abu Dhabi, United Arab Emirates}\\
{\small\textsuperscript{2}Computer Engineering Department, National University of Sciences \& Technology, Pakistan}\\
{\tt\small abdul.basit@nyu.edu, \tt\small khussain39@ce.ceme.edu.pk, 
\tt\small mh6117@nyu.edu, 
\tt\small muhammad.shafique@nyu.edu}}

\renewcommand{\shortauthors}{Abdul Basit et al.}

\begin{abstract}
Large language models (LLMs) are revolutionizing various domains with their remarkable natural language processing (NLP) abilities. However, deploying LLMs in resource-constrained edge computing and embedded systems presents significant challenges. Another challenge lies in delivering medical assistance in remote areas with limited healthcare facilities and infrastructure. To address this, we introduce MedAide, an on-premise healthcare chatbot. It leverages tiny-LLMs integrated with LangChain, providing efficient edge-based preliminary medical diagnostics and support. MedAide employs model optimizations for minimal memory footprint and latency on embedded edge devices without server infrastructure. The training process is optimized using low-rank adaptation (LoRA). Additionally, the model is trained on diverse medical datasets, employing reinforcement learning from human feedback (RLHF) to enhance its domain-specific capabilities. The system is implemented on various consumer GPUs and Nvidia Jetson development board. MedAide achieves 77\% accuracy in medical consultations and scores 56 in USMLE benchmark, enabling an energy-efficient healthcare assistance platform that alleviates privacy concerns due to edge-based deployment, thereby empowering the community.
\end{abstract}

\maketitle

\section{Introduction}
In recent years, large language models (LLMs) \cite{Zhao2023} have emerged as a groundbreaking technology with the potential to revolutionize multiple domains, and are exceedingly gaining research interest. State-of-the-art LLMs, such as GPT-3 \cite{Brown2020} and its successor GPT-4~\cite{Bubeck2023} have revolutionized NLP tasks. Other LLMs are used for various generative tasks like robotics and smart homes/cities. These auto-regressive models learn the statistical patterns and semantic structures of human language. With their vast knowledge base, LLMs can generate coherent and contextually relevant text and understand complex queries. This prowess makes them a valuable resource for a wide range of applications including medicine and autonomous cognitive systems \cite{10.3389/frai.2023.1169595}\cite{jeblick2022chatgpt} and serve as a driving force behind our research. 

Language and interaction are fundamental elements in medicine, enabling healthcare professionals to gather comprehensive insights \cite{Franz2018}, developing a deeper understanding of the individual's health status. Navigating the landscape of medical applications, there is a compelling need for remote assistance as highlighted in Figure 1. A World Bank and WHO report \cite{who} states that half of the world's population lacks essential health services, with healthcare costs pushing 100 million into extreme poverty annually. This underscores the urgent need for universal health coverage.

\begin{figure}[ht]
    \centering
    \includegraphics[width=1\linewidth]{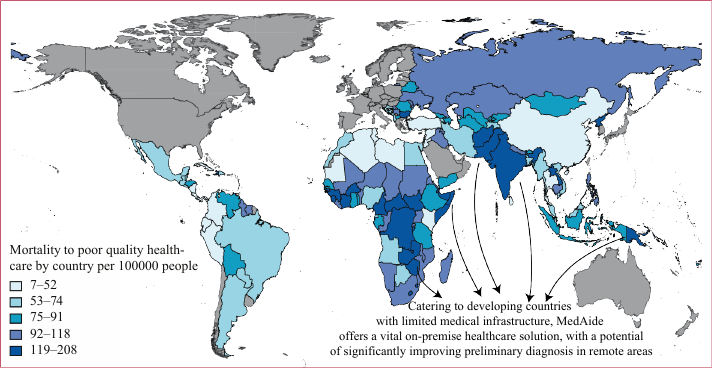}
    \caption{Mortality rate for 137 countries due to inadequate healthcare facilities \cite{Kruk2018MortalityDT} where MedAide could be effective.}
    \label{fig:motivation}    
\end{figure}
Moreover, the deployment of LLMs on resource-constrained devices while minimizing latency presents a significant challenge, see Figure 2. State-of-the-art LLMs \cite{Brown2020} \cite{Bubeck2023} \cite{yin2023survey} heavily rely on server-based deployment, limiting accessibility, and require significant power consumption and computational resources \cite{Zhao2023}. 

\begin{figure}[ht]
    \centering
        \includegraphics[width=1\linewidth]{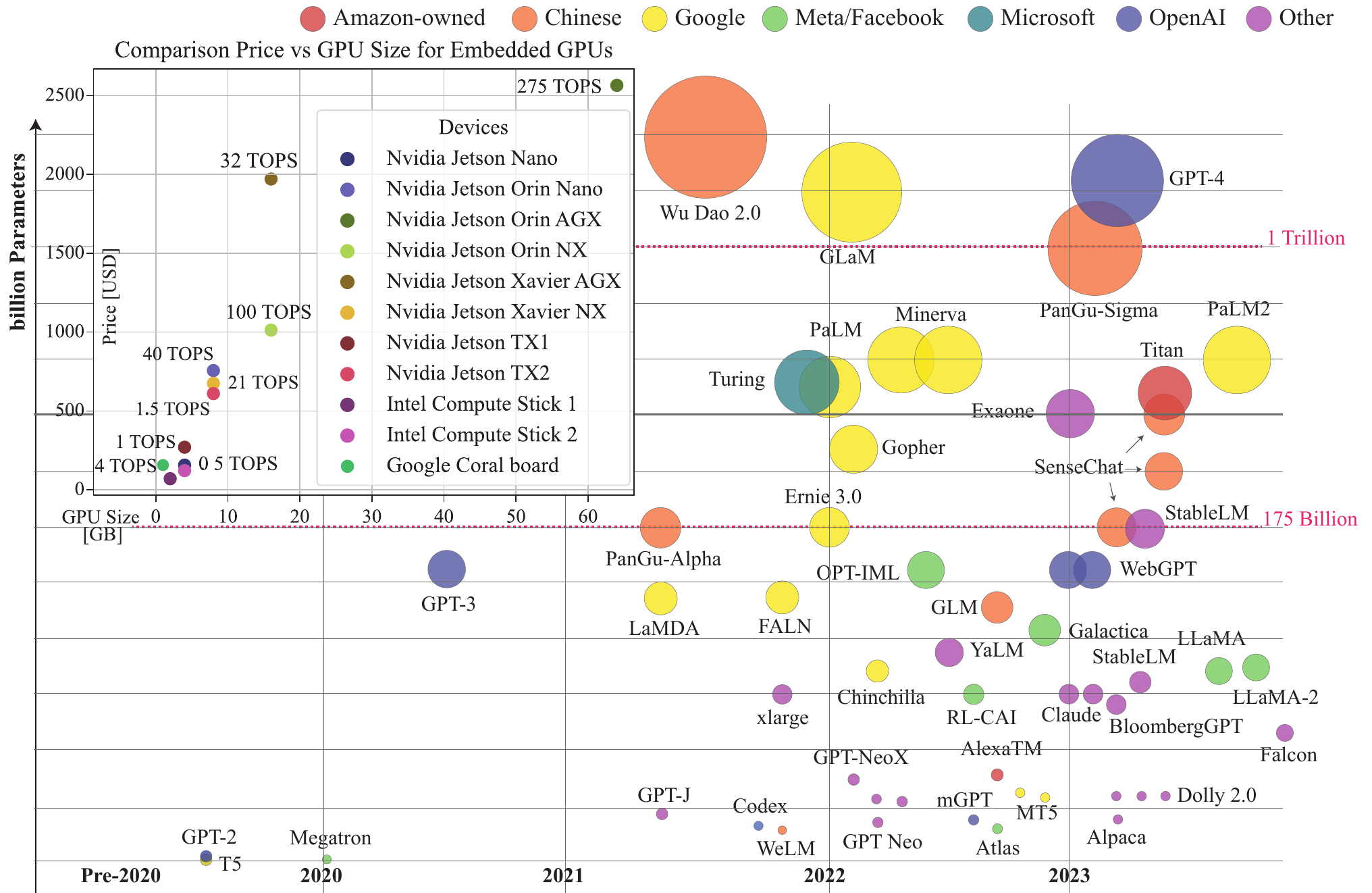}
    \caption{The analysis of embedded devices from NVIDIA, Intel, and Google reveals a significant gap between the capabilities of embedded boards and the requirements of recent LLMs due to escalating memory demands and computational requirements.}
    \label{fig:LLM_progress}    
\end{figure}

\begin{figure*}[ht]
    \centering
    \includegraphics[width=1\textwidth]{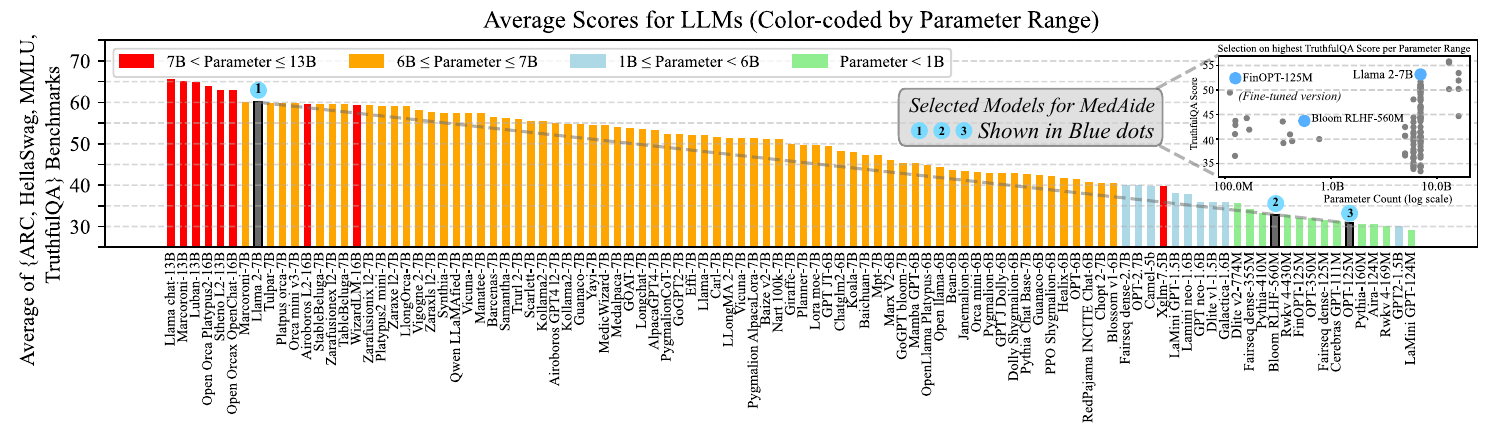}  
    \caption{Open-LLM Leader-board benchmark \cite{Open-LLM-Leaderboard-Report-2023} competing various state-of-the-art LLMs across diverse benchmarks, encompassing TruthfulQA \cite{lin2022truthfulqa}, MMLU \cite{hendrycks2021measuring}, ARC \cite{clark2018think}, and HellaSwag \cite{zellers2019hellaswag} for a comprehensive evaluation.}
    \label{fig:Models}
\end{figure*}

The selection of LLM for edge deployment depends on network parameters, performance metrics, open-source availability, and hardware constraints. We utilized the open-LLM leader-board \cite{Open-LLM-Leaderboard-Report-2023}, which serves as a valuable platform for comprehensive analysis of LLMs. An analysis was conducted to identify suitable LLMs, and among them, LLaMa2-7B\cite{touvron2023llama} emerged as the standout choice owing to its exceptional performance across benchmarks. Moreover, Bloom-560M model \cite{BigScience} and OPT-125M \cite{Zhang2022} models were selected  based on fewer parameter facilitating deployment on resource-constrained embedded devices, open-source availability, and highest TruthfulQA score\cite{lin2022truthfulqa} for their parameter range illustrated in Figure. 3.

We conducted additional analysis to benchmark our models against those with significantly larger parameter counts, which are specialized for medical datasets. Our analysis reveals that OPT-125M and Bloom-560M, with a small memory footprint, exhibit modest accuracy scores of 27.6\%, 29.5\%. These models are particularly attractive due to their relatively lower parameter counts, facilitating easier deployment, thereby achieving low-latency performance. Respectively, LLaMa2-7B achieves 51.9\% accuracy on these tests, reflecting its efficacy in medical applications. Figure 4 showcasing their potential for deployment in medical applications. 
\begin{figure}[ht]
    \includegraphics[width=1\linewidth]{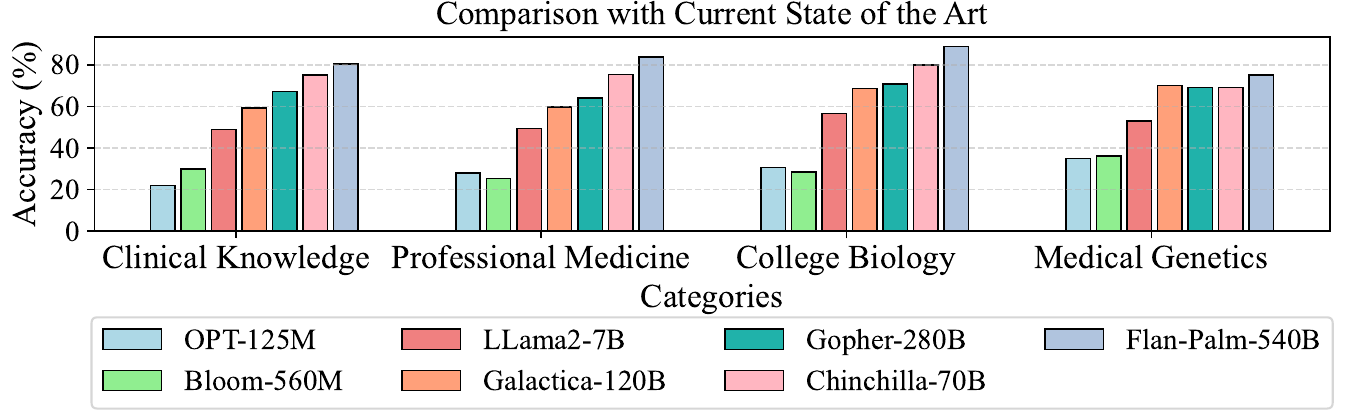}
    \caption{A comprehensive comparison between selected LLMs (OPT, LLaMa2, and Bloom) and state-of-the-art LLMs (Galactica, Gopher, Chinchilla, and Flan-Palm) to evaluate the performance of these models on different domains, shedding light on their feasibility for medical assistance.}
    \label{fig:comparison}    
\end{figure}

LangChain \cite{langchain} plays a crucial role in our model deployment tool-chain through collection of medical databases, optimizing the results through Facebook AI Similarity Search (FAISS) \cite{johnson2019billion}. LangChain efficiently searches a database, to accelerate the retrieval of medical prescription. This streamlined process enhances the overall USMLE score by 2.5 \%, facilitating more effective medical knowledge retrieval and utilization.

To enhance the domain-specific understanding of our models, we curated a custom dataset detailed in Table 1, from various sources including online medical forums, publicly available biomedical databases, and synthesized real-world clinical case studies enabling our models to grasp medical terminologies effectively. By training our models on this specialized dataset, we enrich their knowledge representation and enhance their performance in medical consultation scenarios. Moreover, we devise an experimental validation setup for our models primarily focusing on the doctor-patient interaction aspect which is lacking in most of the available datasets. In this evaluation, chatGPT4 reviews interactive dialogues for authenticity, response effectiveness, and overall performance, ranking MedAide (LLaMa-2) with a 77\% accuracy in comparative analysis.
\begin{figure}[ht]
    \centering
    \includegraphics[width=1\linewidth]{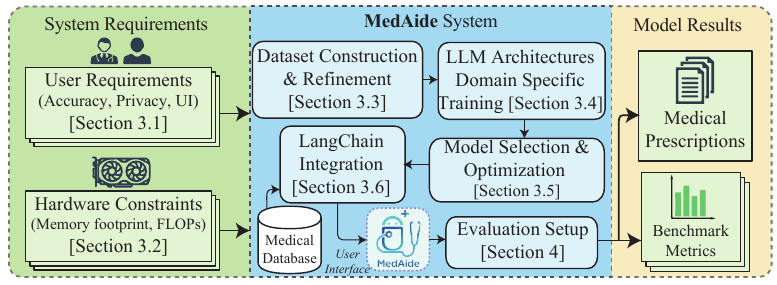}
     \caption{MedAide System overview with the input system requirements and system processes to generate the outputs.}
    \label{fig:introduction}    
\end{figure}

To address the above-discussed research challenges, we present our novel MedAide system with the following contributions:
% \begin{wrapfigure}{r}{0.3\textwidth} % Adjust "-5mm" for overhang into the margin

\begin{itemize}
\item MedAide system (overview in Figure 5) enables on-premise healthcare diagnosis by learning from domain specific dataset, by automatically collecting, refining and updating samples.
\item We investigated MedAide system with a backbone of three prominent LLM architectures that have demonstrated outstanding performance on the Open-LLM leader-board \cite{Open-LLM-Leaderboard-Report-2023}. Our analysis is centered around benchmarking their performance, with a particular emphasis on their effectiveness in the TruthfulQA task, by mitigating false answers learned from imitating human texts.
\item Our proposed MedAide system leverages LLMs with optimizations, enabling seamless deployment on devices such as Nvidia Jetson or consumer-grade GPUs, by identifying hardware constrains through a rigorous model selection criteria for a specific edge device.
\item Towards a practical solution, our system integrates LangChain \cite{langchain} to construct toolchains for effective searching of medical databases for accurate prescription and consultation and medical recommendations.
\end{itemize}

\begin{figure*}[ht]
    \centering
    \includegraphics[width=1\textwidth]{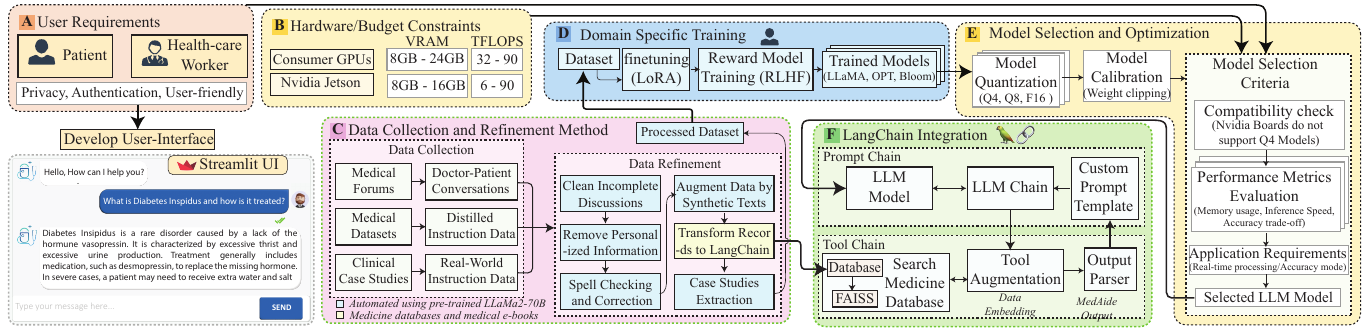}  
    % \caption{An Overview of Our MedAide system: We consider the system’ quality requirements and hardware constraints as inputs to our system. By conducting domain specific training on a refined dataset attained from diverse medical corpus, the resulting LLMs are quantized and the model selection criteria is constructed to deploy them using LangChain empowered by FAISS.}
    \caption{An Overview of Our MedAide: Our system integrates quality requirements and hardware constraints to guide its operation. We employ domain-specific training on a diverse medical dataset to fine-tune LLMs, followed by model quantization. We establish a model selection criteria, ensuring efficient deployment, by integrating LangChain empowered by FAISS.}
    \label{fig:Methodology}
\end{figure*}

\textbf{Paper Organization:} Section 2 discusses state-of-the-art. Section 3 presents the MedAide system in further detail. Section 4 describes the evaluation framework, and results are presented in Section 5. We conclude in Section 6.

\section{Background and Related Work}
LLMs are typically based on transformer architectures \cite{Vaswani2017} \cite{Wolf2019TransformersSN}, which have become the de facto standard for NLP tasks. Transformers leverage attention mechanisms to proficiently capture contextual and long-range dependencies within textual data \cite{Peters2018DeepCW}, while understanding the chain of discussion. LLMs have made significant contributions in various specialized domains. In healthcare, LLMs have been used for medical diagnosis, clinical decision support, and biomedical text mining. Based on a recent study by Dave et al. (2023) \cite{10.3389/frai.2023.1169595}, it is evident that ChatGPT and even GPT-4 demonstrate comparatively lower performance in vertical domains, particularly in the field of medicine. This can be attributed, in part, to the potential insufficiency of medical knowledge among annotators. Consequently, there are substantial prospects for further investigation and enhancement within this domain.

Efforts to overcome this limitation and enhance performance in medical domains hold significant promise for advancing the capabilities of language models in healthcare. ChatDoctor \cite{li2023chatdoctor}, an advanced language model based on the LLaMa model \cite{Touvron2023}, is specifically designed for medical assistance. It simulates doctor-patient conversations, enabling patients to receive accurate diagnoses, personalized medical advice, and appropriate treatment options. Luo et al. \cite{Luo2022} presented BioGPT, a domain-specific generative transformer language model trained on extensive biomedical literature offering generation capabilities tailored specifically for the biomedical domain and achieves superior performance across various biomedical datasets. DrBERT \cite{labrak2023drbert}, a specialized pre-trained language models (PLMs) for the French medical domain, trained on BERT \cite{Devlin2018} introduced a novel approach by leveraging both left and right context during pre-training, resulting in state-of-the-art performance across various medical tasks. HuatuoGPT \cite{huatuogpt-2023}, a distilled language model for medical consultation, trained using ChatGPT synthesised dialogue and real-world data from doctors, employing RLAIF, achieves state-of-the-art performance in medical consultation. These model have excessive server dependence for deployment, due to high parameter count, constraining the availability on GPUs. Moreover, these LLMs often face a challenge known as "hallucination," \cite{hallucination} where they generate plausible but factually incorrect or nonsensical information. This occurs due to their reliance on patterns in the data they were trained on, without an intrinsic understanding of truth. LangChain integrated with LLMs addresses this issue by structuring interactions with the model in a way that mitigates hallucinations.

DoctorGLM \cite{xiong2023doctorglm}, trained using ChatGLM-6B \cite{du2022glm} sets a benchmark on Chinese datasets; however, our work outperforms it with a notable 21\% increase in overall accuracy. Han et al.  \cite{Han2023} proposed MedAlpaca addressed the need for open-source models that prioritize patient privacy by developing specialized dataset for medical applications, highlighting the use-case in medical education. However, MedAide outperforms this work by a considerable margin, as demonstrated in our experimental evaluation section. 

\section{MedAide: An LLM-based System for medical assistance}
In this section we will describe the MedAide system in detail, along with the dataset generation, refinement and training workflow. We elaborate on LLM quantization and model selection along with LangChain integration (see an overview in Figure 6).  
\subsection{User Requirements}
User requirements, pivotal in designing medical applications, dictate both the user interface design and model selection criteria, ensuring alignment with healthcare professionals' and patients' needs. Our analysis highlighted key demands: an intuitive user interface, accurate medical results, secure authentication, and adherence to data privacy laws like HIPAA. Accordingly, we crafted a user interface facilitating secure login and local data storage, overseen by authorized personnel. Additionally, our model selection caters to real-time processing and accuracy modes, tailoring the application to meet these critical user expectations effectively.
\subsection{Hardware Constraints}
There is a trade-off between ensuring high-quality performance and a set of hardware constraints for deployment in consumer-grade GPUs and embedded devices as highlighted in Figure 2. These constraints ensure that LLMs remain within the computational and memory capabilities of the target platforms. The metrics such as storage overhead, based on network parameters, and floating-point operations, quantifying the execution time of the LLMs on GPUs are employed. Given the resource constraints, the trade-off is considering model size for hardware specifications.
\subsection{Dataset Construction \& Refinement}
To prepare our dataset, we conducted an extensive survey of online forums and discussions where individuals post information about their symptoms and seek professional medical advice. We identified several notable forums for data collection, including WebMD, AskDocs, HealthcareMagic, eHealthForum, Icliniq, and HealthTap offering a diverse range of discussions and medical queries and potential to provide data without introducing inherent biases. Additionally, we gathered data from several medical databases for medicine, and clinical records. These online resources provided a significant contribution towards our model training.
% To prepare our dataset, we conducted an extensive survey of online forums and discussions where individuals post information about their symptoms and seek professional medical advice. We identified several notable forums for data collection, including WebMD \cite{WebMd1}, AskDocs \cite{Gomes20202}, HealthcareMagic \cite{healthcaremagic1}, eHealthForum \cite{eHealth}, Icliniq \cite{iCliniq}, and HealthTap\cite{healthtap} offering a diverse range of discussions and medical queries and potential to provide data without introducing inherent biases. Additionally, we gathered data from several medical databases for medicine, and clinical records. These online resources provided a significant contribution towards our model training.

 In some datasets, particularly those obtained from platforms like Kaggle, the original format was in csv, featuring columns such as ``Disease,'' ``Symptom,'' ``Reason,'' ``Tests And Procedures,'' and ``Common Medications.'' To convert this csv data into a question-answer format, we performed data augmentation techniques using a high-fidelity pre-trained LLaMA-70B model. Each row of the csv file was transformed into a common English question using the formulation: ``What are the symptoms, reasons, tests and procedures, and common medications for Panic disorder?" Each row was then converted into a common English question-answer pair in the format:
 
\begin{itemize}
\item The ``instruction'' key contained the original question.
\item The ``output'' key contained the doctor's response.
\item An  "input" field for prompt type, and previous discussion.
\end{itemize}
In some cases, we encountered datasets collected in the Chinese language. To address this challenge, we utilized Google Translate APIs to translate the dataset into English. After compiling the dataset, we conducted an exploratory data refinement phase to filter out incomplete questions, deleted questions, and personal details ensuring the quality and reliability of the dataset. The final dataset comprised approximately 400k instruction-output pairs derived from the mentioned sources.
 % \vspace{2pt}
\begin{table}[ht]
    \caption{Dataset Collection from Different Sources}
    \centering
    \begin{tabular}{l c l}
    \toprule
    \textbf{Source} & \textbf{\# Samples} & \textbf{URL} \\
    \midrule
    HealthcareMagic & 112,641 & \url{www.healthcaremagic.com/} \\
    WebMD & 88,207 & \url{www.webmd.com/} \\
    AskDocs & 24,256 & \textit{www.askdocs.com} \\
    iCliniq & 4,651 & \url{www.icliniq.com/} \\
    HealthTap & 3,647 & \url{www.healthtap.com/} \\
    ehealth & 1,710 & \url{medicalforums.omeka.net} \\
    Huato-26M & 85,000 & \textit{https://doi.org/10.48550} \\
    MedQuAD & 47,457 & \url{github.com/abachaa/MedQuAD} \\
    MedMCQa & 25,679 & \textit{https://doi.org/10.48550} \\
    MedQSum & 1000 & \textit{http://doi.org/10.18653} \\
    Medical Cases & 4363 & \url{www.github.com/adahealth} \\
    \bottomrule
    \end{tabular}
    \label{tab:my_label}
\end{table}

\vspace{-7mm}

% \begin{table}[ht]
%     \caption{Dataset Collection from Different Sources}
%     \centering
%     \begin{tabular}{l c c}
%     \toprule
%     \textbf{Source} & \textbf{Number of Samples} & \textbf{Type} \\
%     \midrule
%     HealthcareMagic \cite{healthcaremagic1} & 112,641 & Forum\\
%     WebMD \cite{WebMd1} & 88,207 & Forum \\
%     AskDocs \cite{Gomes20202} & 24,256 & Forum \\
%     iCliniq \cite{iCliniq} & 4,651 & Forum \\
%     HealthTap \cite{healthtap} & 3,647 & Forum \\
%     ehealth \cite{eHealth} & 1,710 & Forum \\    
%     Translated\_Chinese\_QA \cite{li2023huatuo26m} & 85,000 & Dataset\\
%     MedQuAD \cite{medquad} & 47,457 & Dataset \\
%     MedMCQa \cite{pmlr-v174-pal22a} & 25,679 & Dataset \\
%     MedQSum \cite{MeQSum} & 1000 & Dataset \\
%     Medical Case Report\cite{medCase} & 4363 & Case study \\
%     \bottomrule
%     \end{tabular}
% \end{table}

\subsection{Domain Specific Training Methodology}

Training LLMs such as LLaMa 2-7B \cite{Touvron2023} and Bloom-560M \cite{BigScience} on consumer devices is challenging due to resource limitations. We used Low Rank Adaptation (LoRA) \cite{hu2022lora}  to approximate high-dimensional datasets in lower-dimensional spaces, preserving key features. This technique reduced trainable parameters and GPU memory usage significantly. Bloom-560M, optimized for computational efficiency and accuracy, includes self-attention and feed-forward networks. Similarly, we trained OPT-125M, a model with 12 layers and attention heads, suitable for various hardware configurations used by end-users. Additionally, our approach incorporated Reinforcement Learning from Human Feedback (RLHF) to train a reward model. This model played a pivotal role in refining the training process by providing valuable, human-centric insights. We utilized this reward model to iteratively train a Proximal Policy Optimization (PPO) model, leveraging policy-based rewards. This integration of RLHF allowed for a more nuanced and effective training process, as the PPO model dynamically adjusted its learning strategy based on human feedback. This feedback loop ensured that the model's outputs not only adhered to technical accuracy but also aligned closely with practical, real-world applicability and user expectations. By embedding human feedback directly into the training cycle, we achieved a more robust and contextually aware LLM, capable of addressing the nuanced demands of medical applications with greater precision. Our implementation followed a structured three-step process as highlighted in Figure 7.

\begin{figure}[ht]
    \centering
    \includegraphics[width=0.5\textwidth]{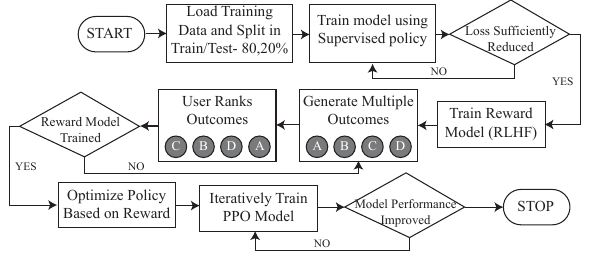}
    \caption{Model Training Workflow.}
    \label{fig:Flowchart}    
\end{figure}

\subsection{Model Selection and Optimization}
Our model selection and optimization process for the LLaMA 2-7B models includes quantization in Q4, Q8, and F16 formats to ensure efficient deployment. Q4 quantization reduces precision to 4 bits, offering smaller model sizes but potentially less accuracy, and is incompatible with Nvidia Jetson boards. Q8 quantization, with 8-bit precision, balances performance and accuracy, while F16 quantization uses 16-bit floating-point representation for complex tasks with reduced size and computational demands. After quantization, the models are calibrated using weight clipping, adjusting weight values to minimize precision loss and maintain performance.

The model selection criteria is designed to ensure efficient deployment on the target hardware. We begin by assessing the compatibility of models with the target hardware, particularly noting that NVIDIA boards do not support Q4 models. This ensures that the chosen model is viable for the intended device. We evaluate metrics such as memory usage, inference speed, and the balance between model accuracy and resource consumption. This evaluation is key to identifying the most efficient model within the hardware's operational constraints. The selection is further refined based on user-defined application requirements, such as the need for real-time processing or a focus on accuracy. This step ensures the model's alignment with the specific functional demands of the application. Finally, we conduct a comparative analysis of various LLMs, weighing the trade-offs each presents. This comprehensive analysis covers the model's feasibility on the device and its conformity to user requirements regarding processing speed and accuracy, ensuring a well-rounded selection process.

In our comprehensive dataset collection, we leveraged medical databases, e-books, and documents for LangChain \cite{langchain} integration. To enhance processing efficiency, we segmented this extensive medical knowledge into 1000-character blocks with a 50-character overlap. Leveraging the robust capabilities of Hugging Face Instruct, we generated embeddings for each block, storing them in our local infrastructure with the support of the FAISS library \cite{johnson2019billion}. Notably, we optimized performance by utilizing the GPU-accelerated version of FAISS, resulting in a remarkable speed boost—achieving search speeds 5-10 x faster than the CPU counterpart.

In response to a user query, we employ FAISS to identify the two closest neighbors using its similarity search function. These data chunks are then presented as contextual input to our prompt chain, in conjunction with the selected LLMs. We significantly alleviate the common issue of hallucination in LLMs. This approach ensures that the medical guidance provided is not only precise but also grounded in reliable and verified medical content, enhancing the overall safety and dependability of the advice offered.

% \begin{figure}[ht]
%     \centering
%     \includegraphics[width=0.45\textwidth]{samples/LangChain.pdf}
%     \caption{LangChain Embedding Generation and Retrieval.}
%     \label{fig:circuit}    
% \end{figure}
% \subsection{Prompt Engineering}
% prompt engineering plays a pivotal role in shaping the interactions between users and MedAide. By carefully crafting prompts or queries, we can guide the system to provide relevant and accurate responses to user inquiries.
% \begin{itemize}
% \item \textbf{Fine-Tune Model Behavior:} We can influence the LLM's responses by designing prompts that explicitly specify the desired information or context. This is particularly useful in tailoring the LLM's output to meet specific application requirements or quality standards.
% \item \textbf{Mitigate Biases:} Through careful prompt construction through LangChain, we can mitigate biases in the LLM's responses. This is critical for addressing ethical considerations and ensuring fair responses.
% \item \textbf{Enhance User Experience:} Well-designed prompts can improve the overall user experience by eliciting more informative and contextually relevant answers from MedAide thereby increasing the system's usability.
% \end{itemize}
\section{Evaluation Setup}
The evaluation setup employs a wide range of software and hardware resources for thorough testing and benchmarking. The software environment integrates various tools such as Python 3.10, Cuda 11.8, PyTorch, Lightning, and other essential libraries, along with FAISS for similarity search. Hardware-wise, the setup features diverse Nvidia GPUs, ranging from high-end models like RTX 6000 Ada to embedded solutions like Jetson AGX Xavier, facilitating a comprehensive evaluation of model performance across different computational environments, as indicated in Figure 8.

The methodology for evaluation involves deploying models on these hardware platforms to assess key metrics such as accuracy, runtime, and benchmark scores. The USMLE and ChatGPT-4 Scores serve as evaluation metrics described later in Section 5.2, providing a standardized measure of model performance. 
% Additionally, the setup includes medical database embedding, allowing us to analyze and output contextually relevant medical information. 

\begin{figure}[ht]
    \centering
    \includegraphics[width=0.48\textwidth]{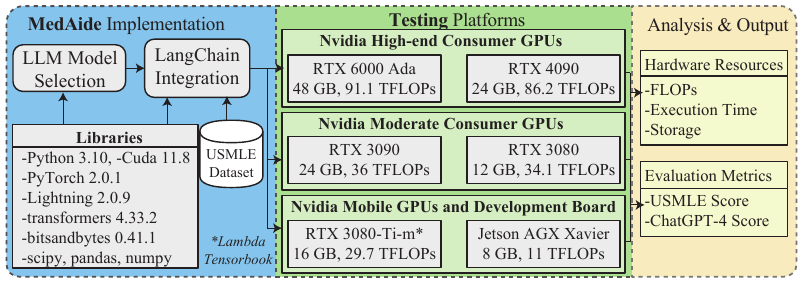}
    \caption{Experimental evaluation setup deployed on Python 3.10 and tested on various testing platforms, and evaluation based on metrics and resources.}
    \label{fig:experimental_setup}    
\end{figure}

\section{Results}
\subsection{Quantitative Analysis of Models}

% We conducted an empirical study and a detailed examination of the model outputs, considering factors such as fluency, coherence, relevance, and domain-specific criteria in medicine. We assessed the outputs from the model for any errors, biases, or limitations, 
In our research, we performed a quantitative analysis of MedAide models using the USMLE QA dataset, a standard for evaluating LLMs in the medical domiain, depicted in Figure 9, also comparing models' performance and GPU memory consumption with SOTA LLMs, highlighting the efficiency. Figure 10 complements this by illustrating latency differences across a range of devices, crucial for practical healthcare applications.
\begin{figure}[ht]
    \centering
    \includegraphics[width=0.5\textwidth]{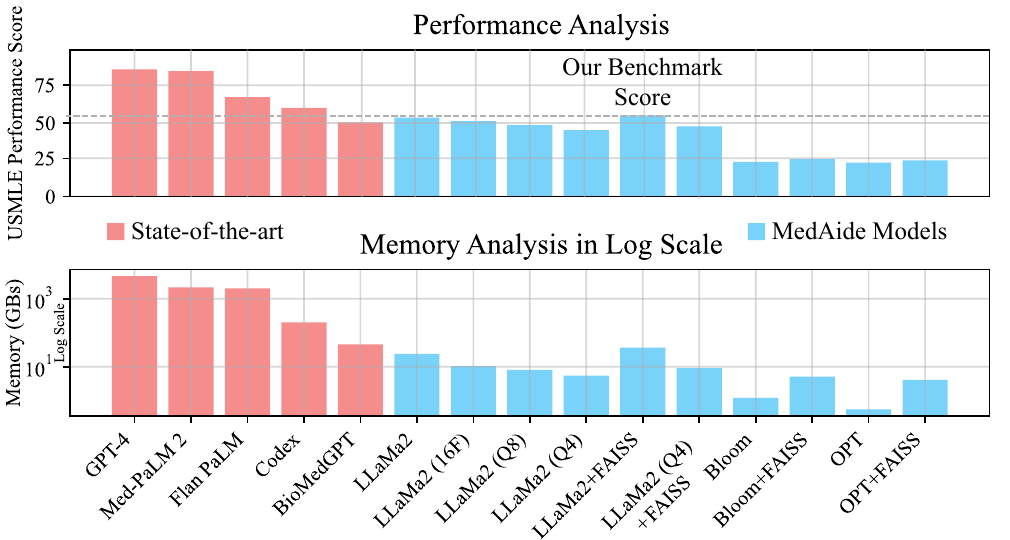}
    \caption{Performance and Memory comparison for different LLM configurations and their respective USMLE scores}
    \label{fig:usmle}    
\end{figure}

\begin{figure}[ht]
    \centering
    \includegraphics[width=0.5\textwidth]{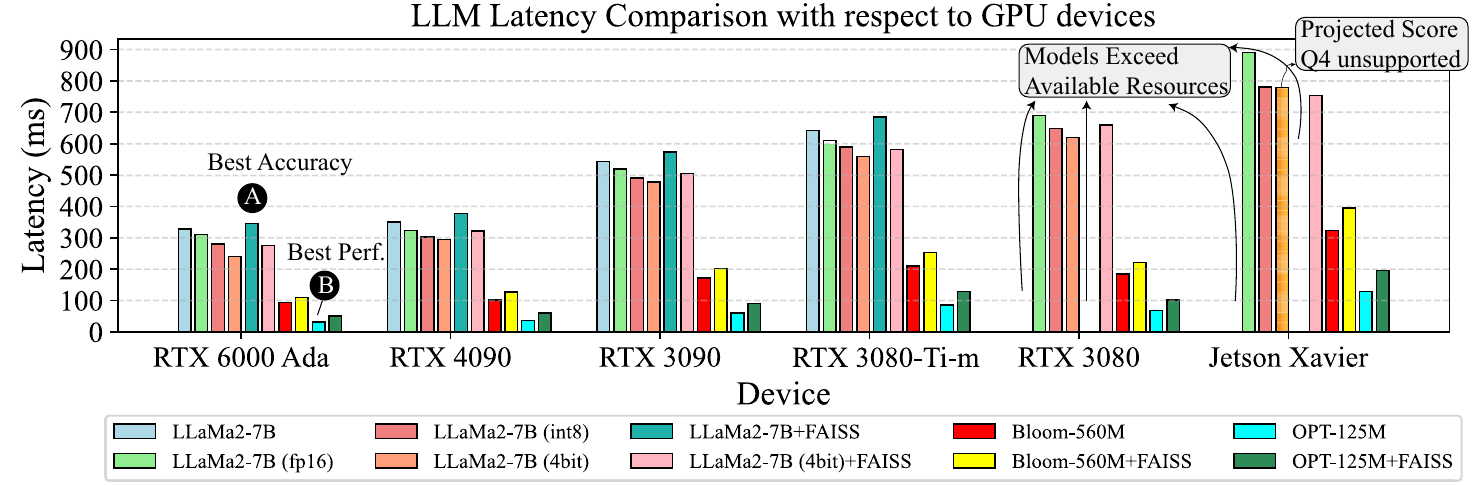}
    \caption{Latency comparison on GPUs architectures, providing a comprehensive overview of model performance across devices.  }
    \label{fig:latency}    
\end{figure}

\subsection{Qualitative Analysis of Models}
This model comparison evaluates the performance of different models by examining their responses to 100 single-turn medical questions. Specifically, the comparison focuses on assessing the ratio of model performance for various categories. The evaluation was conducted using chatGPT4 for reviewing and analyzing the models' outputs, since it scores significantly higher on majority of the benchmarks, considering it as a baseline for performance. The aim of this comparison is to gain insights into how the models perform across various categories and to identify any variations or differences in their effectiveness in health care assessment. Figure 11 presents these findings, and our analysis reveals that our LLaMA 2-7B+FAISS model outperforms MedAlpaca and DoctorGLM by 21\% and 13\%, respectively. This significant margin underscores the efficacy of the LLaMA 2-7B+FAISS model in the evaluated scenarios.

\begin{tcolorbox}[width=1\linewidth, size=small, boxrule=0.45pt,
    before upper={\fontsize{8.15}{9}\selectfont}, left=4pt, right=4pt]
\textbf{Test 1:}
\begin{itemize}
\item \textbf{Instruction:} Suggest treatment for pneumonia
\item \textbf{ChatGPT Output:} The treatment for pneumonia typically involves antibiotics, rest, and supportive care such as staying hydrated and managing symptoms. It is important to consult a healthcare professional for a proper diagnosis.
\item \textbf{MedAide Output:} Hi, it is advisable to take antibiotics like clarithromycin and levofloxacin tablets along with antipyretics like acetaminophen and anti-inflammatory drugs like ibuprofen. It is a good idea to take paracetamol (acetaminophen) tablets to relieve fever.
\end{itemize}
\end{tcolorbox}

\begin{tcolorbox}[width=1\linewidth, size=small, boxrule=0.45pt,
    before upper={\fontsize{8.15}{9}\selectfont}, left=4pt, right=4pt]
\textbf{Test 2:}
\begin{itemize}
  \item \textbf{Instruction:} Is it okay to give Benadryl to a patient with asthma?
  \item \textbf{ChatGPT Output:} The use of Benadryl (diphenhydramine) in patients with asthma should be approached with caution. While Benadryl is an antihistamine that can help alleviate allergic symptoms such as itching, sneezing, and runny nose, it can also cause side effects that may affect individuals with asthma.
  \item \textbf{MedAide Output:} Benadryl treats allergies and it works for asthma as well. It also treats hay fever and many other allergic conditions. However, I would not recommend it as the sole treatment for asthma as it does not address the underlying causes and may not provide sufficient control of asthma symptoms. It is important to consult a healthcare professional for appropriate management of asthma.
\end{itemize}
\end{tcolorbox}

\begin{tcolorbox}[width=1\linewidth, size=small, boxrule=0.5pt,
    before upper={\fontsize{8.15}{9}\selectfont}, left=4pt, right=4pt]
\textbf{Test 3:}
\begin{itemize}
  \item \textbf{Instruction:} I have a tight and painful chest with a dry cough, no fever and no headaches. Could it possibly be coronavirus?
  \item \textbf{ChatGPT Output:} While chest tightness, painful cough, and absence of fever/headaches can be symptoms associated with COVID-19, it is important not to jump to conclusions. These symptoms can also be caused by other respiratory conditions such as bronchitis or asthma. If you are concerned about COVID-19, it is advisable to get tested and consult a healthcare professional for an accurate diagnosis and appropriate guidance.
  \item \textbf{MedAide Output:} Hi dear, since you mentioned you do not have any fever, so you need not worry about coronavirus. But since you have painful chest with cough, you need to go for X-ray of the chest, which will help us to rule out bronchitis. You can also get a routine blood test to rule out any other issues like anemia or liver dysfunction. Regarding the medicines, you can take one course of antibiotics like tablet Cefixime.
\end{itemize}
\end{tcolorbox}

\begin{tcolorbox}[width=1\linewidth, size=small, boxrule=0.5pt,
    before upper={\fontsize{8.15}{9}\selectfont}, left=4pt, right=4pt]
\textbf{Test 4:}
\begin{itemize}
  \item \textbf{Instruction:} What are the fetal effects of maternal von Willebrand disease?
  \item \textbf{ChatGPT Output:} In general, von Willebrand disease (vWD) is a bleeding disorder that affects blood clotting. When it comes to the fetal effects of maternal von Willebrand disease, there can be some potential implications.
  \item \textbf{MedAide Output:} Weak hemostasis in the fetus can lead to intrauterine growth restriction, anemia, pre-term labor and birth, and fetal death.
\end{itemize}
\end{tcolorbox}
\begin{figure}[ht]
    \centering
    \includegraphics[width=0.43\textwidth]{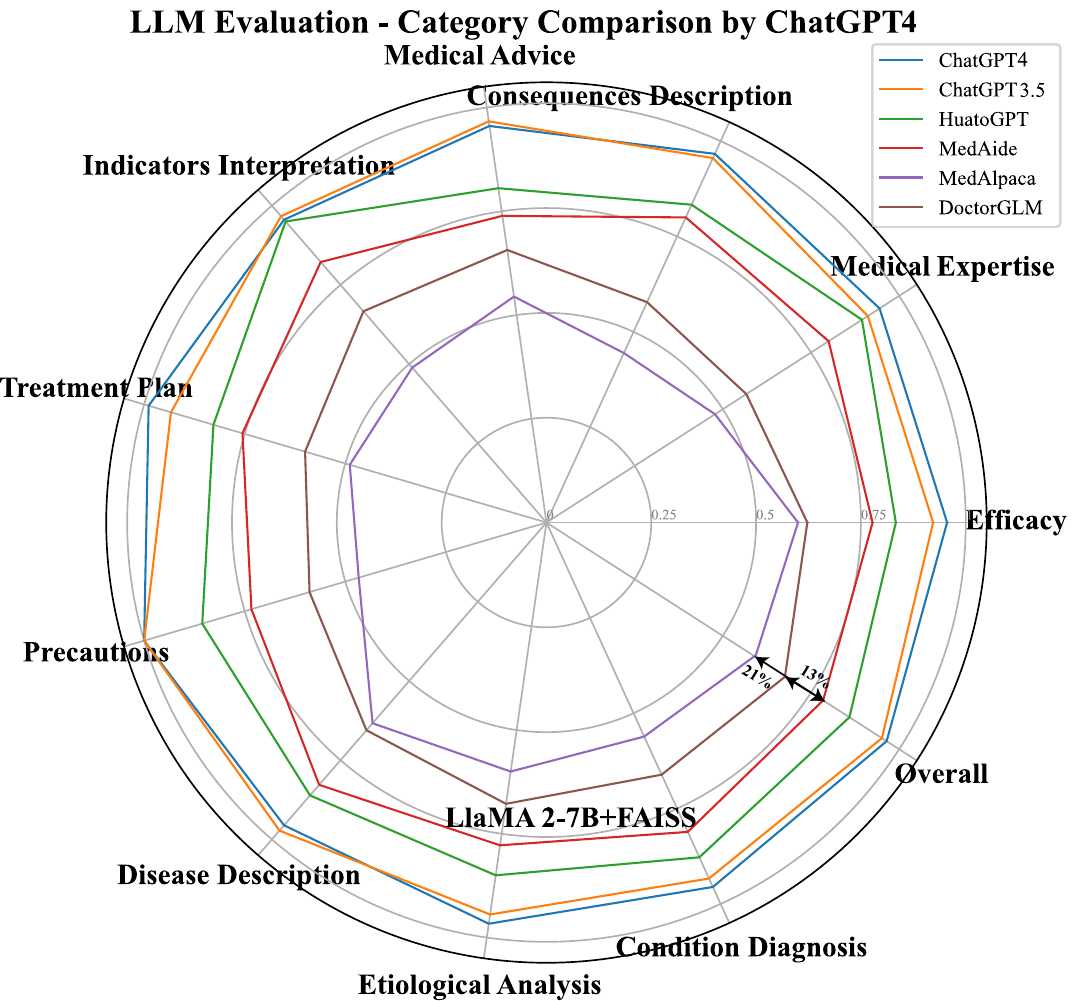}
    \caption{Model perf. ratio for each category evaluated on 100 single-turn questions, reviewed by ChatGPT4.}
    \label{fig:spider}    
\end{figure}

\section{Conclusion}
This paper presented MedAide, an on-premise medical assistant powered by LLMs. MedAide utilizes specialized medical dataset to provide accurate and reliable healthcare support, including answering medical queries, offering personalized recommendations, and aiding in diagnostics. The results demonstrated the effectiveness of MedAide in various medical domains, showcasing its ability to comprehend complex medical queries, generate informative responses, and assist in clinical decision-making. The results showcase its potential to improve medical workflows, diagnostics, and patient care. Integrating MedAide into real-world healthcare settings holds great promise in enhancing the delivery of healthcare services.

\footnotesize  \small %or \scriptsize
% \bibliofont{9}
\bibliographystyle{unsrt}
\bibliography{cite}
% {\footnotesize
% \bibliography{cite}}

\end{document}